\documentclass[10pt,twocolumn,letterpaper]{article}

\usepackage{iccv}
\usepackage{times}
\usepackage{epsfig}
\usepackage{graphicx}
\usepackage{amsmath}
\usepackage{amssymb}
\usepackage{gensymb}
\usepackage{textcomp}
\usepackage{multirow}
\usepackage{array}
\usepackage{booktabs}
\usepackage{paralist}

\usepackage[pagebackref=true,breaklinks=true,letterpaper=true,colorlinks,bookmarks=false]{hyperref}

\iccvfinalcopy 

\def\iccvPaperID{1694} 

\ificcvfinal\fi
\begin{document}

\title{Multi-view Convolutional Neural Networks for 3D Shape Recognition}

\author{
Hang Su \qquad Subhransu Maji \qquad Evangelos Kalogerakis \qquad Erik Learned-Miller\\
University of Massachusetts, Amherst \\
\small{\url{{hsu, smaji, kalo, elm}@cs.umass.edu}}
}

\maketitle

\def\erik#1{\textcolor{blue}{({Erik says: }{#1})}}
\def\sub#1{\textcolor{red}{({Subhransu says: }{#1})}}
\def\vangelis#1{\textcolor{magenta}{({Vangelis says: }{#1})}}
\def\hang#1{\textcolor{cyan}{({Hang says: }{#1})}}

\begin{abstract}
A longstanding question in computer vision concerns the representation
of 3D shapes for recognition: should 3D shapes be represented with 
descriptors operating on their native 3D formats, such as voxel grid
or polygon mesh, or can they be effectively represented with view-based descriptors? 
We address this question in the context of learning to recognize 3D shapes
from a collection of their rendered views on 2D images. 
We first present a standard CNN architecture trained to recognize 
the shapes' rendered views independently of each other, 
and show that a 3D shape can be recognized even from a single view 
at an accuracy far higher than using state-of-the-art 
3D shape descriptors. Recognition rates further 
increase when multiple views of the shapes are provided. 
In addition, we present a novel CNN architecture that combines 
information from multiple views of a 3D shape into a single and
compact shape descriptor offering even better recognition performance. 
The same architecture can be applied
to accurately recognize human hand-drawn sketches of shapes.
We conclude that a collection of 2D views can
be highly informative for 3D shape recognition and is amenable
to emerging CNN architectures and their derivatives.
\end{abstract}

\section{Introduction}

\begin{figure*}[t]
\begin{center}
   \includegraphics[width=0.8\linewidth,trim=0pt 3pt 0pt 0pt,clip=true]{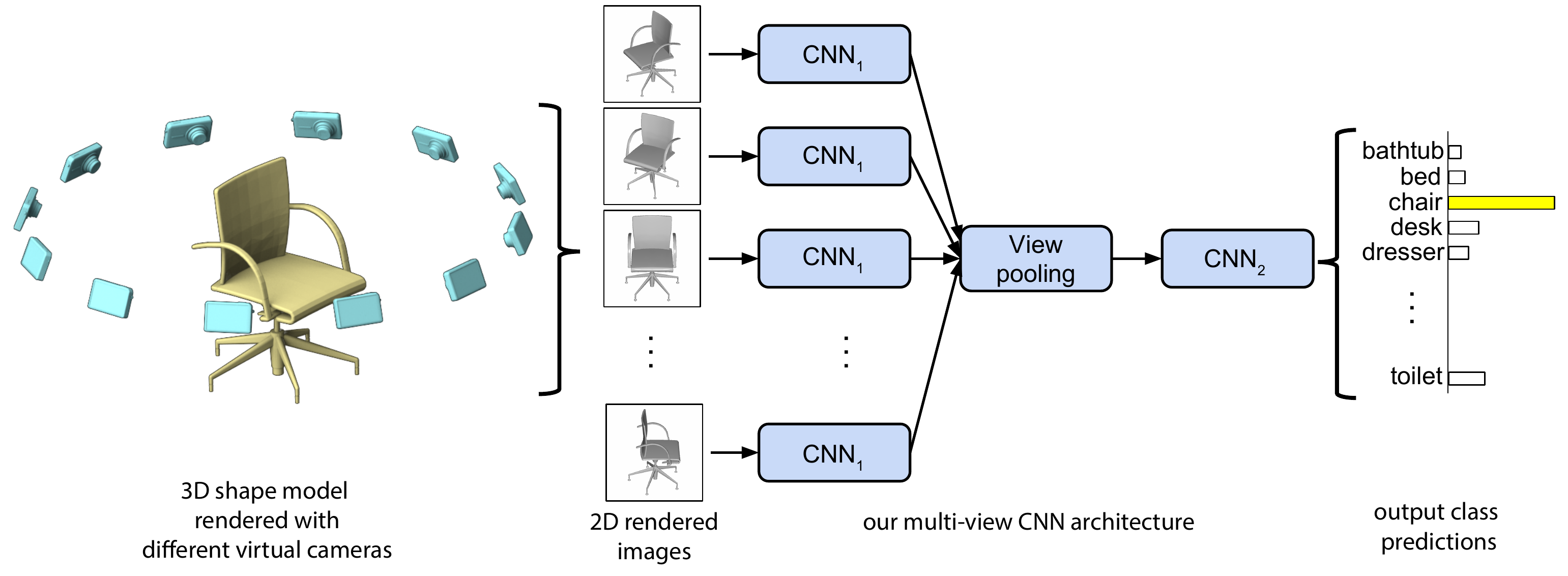}
\end{center}
\vskip -3mm
   \caption{Multi-view CNN for 3D shape recognition (illustrated using the 1\textsuperscript{st} camera setup). At test time a 3D shape is rendered from 12 different views and are passed thorough CNN$_1$ to extract view based features. These are then pooled across views and passed through CNN$_2$ to obtain a compact shape descriptor.}
\label{fig:network}
\vskip -3mm
\end{figure*}

One of the fundamental challenges of computer vision is to draw
inferences about the three-dimensional (3D) world from two-dimensional
(2D) images. Since one seldom has access to 3D object models, one must
usually learn to recognize and reason about 3D objects based upon
their 2D appearances from various viewpoints. Thus, computer vision
researchers have typically developed object recognition algorithms
from 2D features of 2D images, and used them to classify new 2D pictures
of those objects.

But what if one does have access to 3D models of each object of
interest?  In this case, one can directly train recognition algorithms
on 3D features such as voxel occupancy or surface curvature. 
The possibility of building such classifiers
of 3D shapes directly from 3D representations has recently emerged due
to the introduction of large 3D shape repositories, such as 3D 
Warehouse, TurboSquid, and Shapeways. 
For example, when Wu et~al.~\cite{Wu:2015:ModelNet}
introduced the ModelNet 3D shape database, they presented a
classifier for 3D shapes using a deep belief network architecture
trained on voxel representations. 

While intuitively, it seems logical to build 3D shape classifiers directly
from 3D models, in this paper we present a seemingly counterintuitive 
result -- that by building classifiers of 3D shapes from 2D image renderings 
of those shapes, we can actually {\em dramatically outperform} the classifiers
built directly on the 3D representations. In particular, a convolutional neural network (CNN) trained on
a fixed set of rendered views of a 3D shape and only provided with a \emph{single} view at test time increases category recognition
accuracy by a remarkable 8\% (77\% $\rightarrow$ 85\%) over the best models~\cite{Wu:2015:ModelNet} trained on 3D representations.
With more views  provided at test time, its performance further increases. \ 

One reason for this result is the relative efficiency of the 2D versus the 3D representations. In particular, while a
full resolution 3D representation contains all of the information
about an object, in order to use a voxel-based representation in a
deep network that can be trained with available samples
and in a reasonable amount of time, it would appear that the
resolution needs to be significantly reduced. For example, 3D ShapeNets
use a coarse representation of shape, a 30$\times$30$\times$30 grid of binary voxels. In contrast a single projection of the 3D model of the same input size corresponds to an image of 164$\times$164 pixels, or slightly smaller if multiple projections are used. Indeed, there is an inherent trade-off between increasing the amount of explicit depth information (3D models) and increasing spatial resolution (projected 2D models).

Another advantage of using 2D representations is that we can leverage
(i) advances in image descriptors~\cite{lowe99object, perronnin10improving} and (ii) massive image databases (such as ImageNet~\cite{deng09imagenet}) to pre-train our CNN architectures. Because images are ubiquitous and large labeled 
datasets are abundant, we can learn a good deal about generic features for 
2D image categorization and then fine-tune to specifics about 3D model
projections. While it is possible that some day as much 3D training
data will be available, for the time being this is a significant advantage
of our representation.

Although the simple strategy of classifying views independently works remarkably well (Sect.~\ref{sec:proto1}), we present new ideas for how to ``compile" the information in multiple 2D views of an object into a compact object descriptor using a new architecture called \emph{multi-view CNN} (Fig.~\ref{fig:network} and Sect.~\ref{sec:proto2}). This descriptor is at least as informative for classification (and for retrieval is slightly more informative) than the full collection of view-based descriptors of the object. Moreover it facilitates efficient retrieval using either a similar 3D object or a simple hand-drawn sketch, without resorting to slower methods that are based on pairwise comparisons of image descriptors. We present state-of-the-art results on 3D object classification, 3D object retrieval using 3D objects, and 3D object retrieval using sketches (Sect.~\ref{sec:results}).

Our multi-view CNN is related to ``jittering" where transformed copies of the data are added during training to learn invariances to transformations such as rotation or translation. In the context of 3D recognition the views can be seen as jittered copies. The multi-view CNN learns to combine the views instead of averaging, and thus can use the more informative views of the object for prediction while ignoring others. Our experiments show that this improves performance (Sect.~\ref{sec:result-3d}) and also lets us visualize informative views of the object by back-propagating the gradients of the network to the views (Fig.~\ref{fig:saliency}). Even on traditional image classification tasks multi-view CNN can be a better alternative to jittering. For example, on the sketch recognition benchmark \cite{Eitz:2012:HDH} a multi-view CNN trained on jittered copies performs better than a standard CNN trained with the same jittered copies (Sect.~\ref{sec:sketch}). 

Pre-trained CNN models, data, and the complete source code to reproduce the results in the paper are available at \url{http://vis-www.cs.umass.edu/mvcnn}.

\section{Related Work}
\label{sec:related}

Our method is related to prior work on shape descriptors for 3D objects and image-based CNNs. Next we discuss representative work in these areas. 

\vskip -2mm
\paragraph{Shape descriptors.} A large corpus of shape descriptors has been developed for drawing inferences about 3D objects in both the computer vision and graphics literature. Shape descriptors can be classified into two broad categories: \emph{3D shape descriptors} that directly work on the native 3D representations of objects, such as polygon meshes, voxel-based discretizations, point clouds, or implicit surfaces, and \emph{view-based descriptors} that describe the shape of a 3D object by ``how it looks'' in a collection of 2D projections. 

With the exception of the recent work of Wu et~al.~\cite{Wu:2015:ModelNet} which learns shape descriptors from the voxel-based representation of an object through 3D convolutional nets, previous 3D shape descriptors were largely ``hand-designed'' according to a particular geometric property of the shape surface or volume. For example, shapes can be represented with histograms or bag-of-features models constructed out of surface normals and curvatures \cite{Horn:1984:EGI}, distances, angles, triangle areas or tetrahedra volumes gathered at sampled surface points \cite{Osada:2002:SD}, properties of spherical functions defined in volumetric grids \cite{Kazhdan:2003:RISH}, local shape diameters measured at densely sampled surface points \cite{Chaudhuri:2010:ddsc}, heat kernel signatures on polygon meshes \cite{Bronstein:2011:SGGW,Bronstein:2012:ISC}, or extensions of the SIFT and SURF feature descriptors to 3D voxel grids \cite{Knopp:2010:HTS}. Developing classifiers and other supervised machine learning algorithms on top of such 3D shape descriptors poses a number of challenges. First, the size of organized databases with annotated 3D models is rather limited compared to image datasets, \eg, ModelNet contains about 150K shapes (its 40 category benchmark contains about 4K shapes). In contrast, the ImageNet database~\cite{deng09imagenet} already includes tens of millions of annotated images.
Second, 3D\ shape descriptors tend to be very high-dimensional, making classifiers
prone to overfitting due to the so-called  `curse of dimensionality'. 


On the other hand view-based descriptors have a number of desirable properties: they are relatively low-dimensional, efficient to evaluate, and robust to 3D shape representation artifacts, such as holes, imperfect polygon mesh tesselations, noisy surfaces. The rendered shape views can also be directly compared with other 2D images, silhouettes or even hand-drawn sketches. An early example of a view-based approach is the work by Murase and Nayar \cite{Murase:1995:VLR} that recognizes objects by matching their appearance in parametric eigenspaces formed by large sets of 2D renderings of 3D models under varying poses and illuminations. Another example, which is particularly popular in computer graphics setups, is the LightField descriptor \cite{Chen:2003:ovsb} that extracts a set of geometric and Fourier descriptors from object silhouettes rendered from several different viewpoints. Alternatively, the silhouette of an object can be decomposed into parts and then represented by a directed acyclic graph (shock graph) \cite{Macrini:2002:VBO}.  Cyr and Kimia \cite{Cyr:2004:SAA} defined similarity metrics based on curve matching and grouped similar views, called aspect graphs of 3D models~\cite{koenderink1976singularities}. Eitz et al.~\cite{Eitz:2012:SSR} compared human sketches with line drawings of 3D models produced from several different views based on local Gabor filters, while Schneider et al.~\cite{Schneider:2014:SCC} proposed using Fisher vectors~\cite{perronnin10improving} on SIFT features~\cite{lowe99object} for representing human sketches of shapes. These descriptors are largely ``hand-engineered'' and some do not generalize well across different domains. 

\vskip -2mm
\paragraph{Convolutional neural networks.} 
Our work is also related to recent advances in image recognition using CNNs~\cite{Krizhevsky:2012:ICD}. In particular CNNs trained on the large datasets such as ImageNet have been shown to learn general purpose image descriptors for a number of vision tasks such as object detection, scene recognition, texture recognition and fine-grained classification~\cite{Donahue:2013:DeCAF,girshick14rich,razavin14cnn-features,cimpoi14describing}. We show that these deep architectures can be adapted to specific domains including shaded illustrations of 3D objects, line drawings, and human sketches to produce descriptors that have superior performance compared to other view-based or 3D shape descriptors in a variety of setups. Furthermore, they are compact and efficient to compute. There has been existing work on recognizing 3D objects with CNNs \cite{lecun04learning} using two concatenated views (binocular images) as  input. Our network instead
learns a shape representation that aggregates information from any number of input views without any specific
ordering, 
and always outputs a compact shape descriptor of the same size. Furthermore,
we
leverage both image and shape datasets to train our network. 
\vskip -2mm
\paragraph{}Although there is significant work on 3D and 2D shape descriptors, and estimating informative views of the objects (or, aspect graphs), there is relatively little work on learning to combine the view-based descriptors for 3D shape recognition. Most methods resort to simple strategies such as performing exhaustive pairwise comparisons of descriptors extracted from different views of each shape, or concatenating descriptors
from ordered, consistent views. In contrast our multi-view CNN architecture learns to recognize 3D shapes from views of the shapes using image-based CNNs but in the \emph{context} of other views via a view-pooling layer. As a result, information from multiple views is effectively accumulated into a single, compact shape descriptor.

\section{Method}

As discussed above, our focus in this paper is on developing view-based
descriptors for 3D shapes that are trainable, produce informative representations
for recognition and retrieval tasks, and are efficient to compute.

Our view-based representations start from multiple views of a 3D shape, generated by a rendering engine. A simple way to use multiple views is to generate a 2D image descriptor per each view, and then use the individual descriptors directly for recognition tasks based on some voting or alignment scheme. For example, a na\"ive approach would be to average the individual descriptors, treating all the views as equally important. Alternatively, if the views are rendered in a reproducible order, one could also concatenate the 2D descriptors of all the views. Unfortunately, aligning a 3D shape to a canonical orientation is hard and sometimes ill-defined. In contrast to the above simple approaches, an aggregated representation combining features from multiple views is more desirable since it yields a single, compact descriptor representing the 3D shape. 

Our approach is to learn to combine information from multiple views using a unified CNN architecture that includes a view-pooling layer (Fig.~\ref{fig:network}). All the parameters of our CNN architecture are learned discriminatively to produce a single compact descriptor for the 3D shape. Compared to exhaustive pairwise comparisons between single-view representations of 3D shapes, our resulting descriptors can be directly used to compare 3D shapes leading to significantly higher computational efficiency.

\subsection{Input: A Multi-view Representation}
\label{sec:rendering}
3D models in online databases are typically stored as polygon meshes, which are collections of points connected with edges forming faces. To generate rendered views of polygon
meshes, we use the Phong reflection model \cite{Phong:1975:ICG}. 
The mesh polygons are rendered under a perspective projection and the pixel color is determined by interpolating the reflected intensity of the polygon vertices. Shapes are uniformly scaled to fit into the viewing volume.

To create a multi-view shape representation, we need to setup viewpoints (virtual cameras) for rendering each mesh. We experimented with two camera setups. For 
{\bf the 1\textsuperscript{st} camera setup}, we assume that the input shapes are upright oriented along a consistent axis (\eg,  z-axis).  Most models in modern online repositories, such as the 3D Warehouse, satisfy this requirement, and some previous recognition methods
also follow the same assumption \cite{Wu:2015:ModelNet}. In this case, we create 12 rendered views by placing 12 virtual cameras around the mesh every 30 degrees (see Fig.~\ref{fig:network}). The cameras are elevated 30 degrees from the ground plane, pointing towards the centroid of the mesh.  The centroid is calculated as the weighted average of the mesh face centers, where the weights are the face areas. For {\bf the 2\textsuperscript{nd} camera setup}, we  do not make use of the assumption about consistent
upright  orientation of shapes. In this case, we render from
several more viewpoints since we do not know beforehand which ones yield
good representative views of the object. The renderings are generated by placing 20 virtual cameras at
the 20 vertices of an icosahedron enclosing the shape. All  cameras point  towards the centroid of the mesh. Then we generate 4
 rendered views from each camera, using 0, 90, 180, 270 degrees rotation along the axis passing through the camera and the object centroid, yielding total 80 views.

We note that using different shading coefficients or illumination models did not affect our output descriptors due to the invariance of the learned filters to illumination changes, as also observed in image-based CNNs \cite{Krizhevsky:2012:ICD,Donahue:2013:DeCAF}. Adding more or different viewpoints is trivial, however, we found that the above camera setups were already enough to achieve high performance. Finally, rendering each mesh from all the viewpoints takes no more than ten milliseconds on modern graphics hardware. 

\subsection{Recognition with Multi-view Representations}
\label{sec:proto1}

We claim that our multi-view representation contains rich information about 3D shapes and 
can be applied to various types of tasks. In the first setting, we make use of existing 
2D image features directly and produce a descriptor for each view. This is the most straightforward 
approach to utilize the multi-view representation. However, it results in multiple 2D image descriptors per 3D shape, one per view, which need to be integrated somehow for recognition tasks. 

\vskip -2mm
\paragraph{Image descriptors.} We consider two types of image descriptors for each 2D view: a state-of-the-art ``hand-crafted" image descriptor based on Fisher vectors \cite{Sanchez:2013:ICF} with multi-scale SIFT, as well as CNN activation features \cite{Donahue:2013:DeCAF}.

The Fisher vector image descriptor is implemented using VLFeat~\cite{vedaldi08vlfeat}. For each image multi-scale SIFT descriptors are extracted densely. These are then  projected to 80 dimensions with PCA, followed by Fisher vector pooling with a Gaussian mixture model with 64 components, square-root and $\ell_2$ normalization.

For our CNN features we use the VGG-M network from~\cite{Chatfield:2014:RDD} which consists of mainly five convolutional layers conv$_{1,\ldots,5}$ followed by three fully connected layers fc$_{6,\ldots,8}$ and a softmax classification layer. The penultimate layer fc$_7$ (after ReLU non-linearity, 4096-dimensional) is used as image descriptor. The network is pre-trained on ImageNet images from 1k categories, and then fine-tuned on all 2D views of the 3D shapes in training set. As we show in our experiments, fine-tuning improves performance significantly. Both Fisher vectors and CNN features yield very good performance in classification and retrieval compared with popular 3D shape descriptors (\eg, SPH \cite{Kazhdan:2003:RISH}, LFD \cite{Chen:2003:ovsb}) as well as 3D ShapeNets \cite{Wu:2015:ModelNet}. 

\vskip -2mm
\paragraph{Classification.} We train one-vs-rest linear SVMs (each view is treated as a separate training sample) 
to classify shapes using their image features. At test time, we simply sum up the
SVM decision values over 
all 12 views and return the class with the highest sum. Alternative approaches, \eg, averaging image descriptors, 
lead to worse accuracy. 

\vskip -2mm
\paragraph{Retrieval.} A distance or similarity measure is required for retrieval tasks. For shape $\mathbf{x}$ 
with $n_x$ image descriptors and shape $\mathbf{y}$ with $n_y$ image descriptors, the distance between them is defined 
in Eq.~\ref{eq:avgmindist}. Note that the distance between two 2D images is defined as the $\ell_2$ distance between their feature vectors, 
\ie $\|\mathbf{x}_i-\mathbf{y}_j\|_2$. 

\begin{align}
\!\! d(\mathbf{x},\mathbf{y}) = \frac{\sum_j \min_i \|\mathbf{x}_i-\mathbf{y}_j\|_2}{2n_y} + 
                                 \frac{\sum_i \min_j \|\mathbf{x}_i-\mathbf{y}_j\|_2}{2n_x} \label{eq:avgmindist}
\end{align}

To interpret this definition, we can first define the distance between a 2D image $\mathbf{x}_i$ and 
a 3D shape $\mathbf{y}$ as $d(\mathbf{x}_i,\mathbf{y})=\min_j \|\mathbf{x}_i-\mathbf{y}_j\|_2$. Then given all $n_x$ 
distances between $\mathbf{x}$'s 2D projections and $\mathbf{y}$, the distance between these two shapes is computed by
 simple averaging. 
In Eq.~\ref{eq:avgmindist}, this idea is applied in both directions to ensure symmetry. 

We investigated alternative distance measures, such as minimun distance 
among all $n_x \cdot n_y$ image pairs and the distance between average image descriptors, but they all led 
to inferior performance. 

\begin{table*}
\setlength{\tabcolsep}{4pt}
\begin{center}
\begin{tabular}{lcccccc}
\toprule
\multicolumn{1}{c}{\multirow{2}{*}{\textbf{Method}}} & \multicolumn{3}{c}{\textbf{Training Config.}} & \textbf{Test Config.} & \multirow{2}{2.1cm}{\centering \textbf{Classification (Accuracy)}}  & \multirow{2}{2.1cm}{\centering \textbf{Retrieval (mAP)}}\\
\cmidrule(lr){2-4} \cmidrule(lr){5-5} & \textbf{Pre-train} & \textbf{Fine-tune} & \textbf{\#Views} & \textbf{\#Views} && \\
\midrule
(1) SPH \cite{Kazhdan:2003:RISH} & - & - & - & - & 68.2\% & 33.3\%\\
(2) LFD \cite{Chen:2003:ovsb} & - & - & - & - & 75.5\% & 40.9\% \\
(3) 3D ShapeNets \cite{Wu:2015:ModelNet} & ModelNet40 & ModelNet40 & - & - & 77.3\% & 49.2\%\\
\midrule
(4) FV & - & ModelNet40 & 12 & 1 & 78.8\% & 37.5\% \\
(5) FV, 12$\times$ & - & ModelNet40 & 12 & 12 & 84.8\% & 43.9\% \\
(6) CNN & ImageNet1K & - & - & 1 & 83.0\% & 44.1\% \\
(7) CNN, f.t. & ImageNet1K & ModelNet40 & 12 & 1 & 85.1\% & 61.7\%\\
(8) CNN, 12$\times$ & ImageNet1K & - & - & 12 & 87.5\% & 49.6\% \\
(9) CNN, f.t.,12$\times$ & ImageNet1K & ModelNet40 & 12 & 12 & 88.6\% & 62.8\% \\
\midrule
(10) MVCNN, 12$\times$ & ImageNet1K & - & - & 12 & 88.1\% & 49.4\% \\
(11) MVCNN, f.t., 12$\times$ & ImageNet1K & ModelNet40 & 12 & 12 & 89.9\% & 70.1\% \\
(12) MVCNN, f.t.$+$metric, 12$\times$ & ImageNet1K & ModelNet40 & 12 & 12 & 89.5\% & \textbf{80.2}\% \\
(13) MVCNN, 80$\times$ & ImageNet1K & - & 80 & 80 & 84.3\% & 36.8\% \\
(14) MVCNN, f.t., 80$\times$ & ImageNet1K & ModelNet40 & 80 & 80 & \textbf{90.1}\% & 70.4\% \\
(15) MVCNN, f.t.$+$metric, 80$\times$ & ImageNet1K & ModelNet40 & 80 & 80 & \textbf{90.1}\% & 79.5\% \\
\bottomrule
\multicolumn{7}{l}{
    \begin{minipage}{12cm}
        \item[*] f.t.$=$fine-tuning, metric$=$low-rank Mahalanobis metric learning
    \end{minipage}
}\\
\end{tabular}
\vskip -3mm
\end{center}
\caption{Classification and retrieval results on the ModelNet40 dataset. On the top are results using state-of-the-art 3D shape descriptors. Our view-based descriptors including Fisher vectors (FV)  significantly outperform these even when a single view is available at test time (\#Views = 1). When multiple views (\#Views=12 or
80) are available at test time, the performance of view-based methods improve significantly. The multi-view CNN (MVCNN) architecture outperforms the view-based methods, especially for retrieval.}
\label{table:modelnet}
\vskip -3mm
\end{table*}

\subsection{Multi-view CNN: Learning to Aggregate Views}
\label{sec:proto2}

Although  having multiple 
separate descriptors for each 3D shape can be successful for classification and retrieval compared to existing 3D descriptors, it can be inconvenient and inefficient in many cases. For example, in 
Eq.~\ref{eq:avgmindist}, we need to compute all $n_x \times n_y$ pairwise distances between images in order to 
compute distance between two 3D shapes. Simply averaging or concatenating the image descriptors leads to inferior 
performance. In this section, we focus on the problem of learning to aggregate multiple views in order to 
synthesize the information from all views into a single, compact 3D shape descriptor. 

We design the multi-view CNN (MVCNN) on top of image-based CNNs (Fig.~\ref{fig:network}). Each image in a 3D shape's multi-view
representation is passed through the first part of the network (CNN$_1$) separately, aggregated at a view-pooling layer,   
and then sent through the remaining part of the network (CNN$_2$). All branches in the first part of the network share the same parameters in CNN$_1$. 
We use element-wise maximum operation across the views in the view-pooling layer. An alternative is element-wise mean operation, but it is not as effective 
in our experiments. 
The view-pooling layer can be placed anywhere in the network. We show in our experiments that it should be placed 
close to the last convolutional layer (conv$_5$) for optimal classification and retrieval performance. 
View-pooling layers are closely related to max-pooling layers and maxout layers \cite{Goodfellow:2013:Maxout}, 
with the only difference being the dimension that their pooling operations are carried out on. 
The MVCNN is a  directed acyclic graphs and can be trained or fine-tuned using stochastic gradient descent with back-propagation.

Using fc$_7$ (after ReLU non-linearity) in an MVCNN as an aggregated
shape descriptor, we achieve higher performance than using separate
image descriptors from an image-based CNN directly, 
especially in retrieval (62.8\% $\rightarrow$ 70.1\%). Perhaps more importantly,
the aggregated descriptor is readily available for a
variety of tasks, \eg, shape classification and retrieval, and offers
significant speed-ups against multiple image descriptors.

An MVCNN can also be used as a general framework to integrate perturbed image samples (also known as data jittering). We illustrate this capability 
of MVCNNs in the context of sketch recognition in Sect.~\ref{sec:sketch}.

\vskip -2mm
\paragraph{Low-rank Mahalanobis metric.} Our MVCNN is fine-tuned for classification, and thus retrieval performance is not directly optimized. Although we could train it with a different objective function suitable for retrieval, we found that a simpler approach can readily yield a significant retrieval performance boost (see row 12 in Tab.~\ref{table:modelnet}). We learn a Mahalanobis metric $W$ that directly projects MVCNN descriptors $\mathbf{\phi} \in \mathbb{R}^d$ to $W \mathbf{\phi} \in \mathbb{R}^p$, such that the $\ell_2$ distances in the projected space are small between shapes of the same category, and large otherwise. We use the large-margin metric learning algorithm and implementation from \cite{Simonyan13FVFW}, with  $p<d$ to make the final descriptor compact ($p=128$ in our experiments). 
The fact that we can readily use metric learning over the output shape descriptor demonstrates another advantage of using MVCNNs. 

\section{Experiments}
\label{sec:results}

\subsection{3D Shape Classification and Retrieval}
\label{sec:result-3d}

We evaluate our shape descriptors on the Princeton ModelNet dataset \cite{Web:ModelNet}. 
ModelNet currently contains 127,915 3D CAD models from 662 categories.\footnote{As of 09/24/2015.} A 40-class well-annotated subset containing 12,311 shapes from 40 common categories, ModelNet40, is provided on the ModelNet website. For our experiments, we use the same training and test split of ModelNet40 as in \cite{Wu:2015:ModelNet}.\footnote{Based
on our correspondence with the authors of \cite{Wu:2015:ModelNet}, for each category the first 80 shapes in the ``train" folder (or all shapes if there are fewer than 80) should be used for training, while the first 20 shapes in the ``test" folder should be used
for testing.}

Our shape descriptors are compared against the 3D ShapeNets by Wu \etal \cite{Wu:2015:ModelNet}, the Spherical Harmonics descriptor (SPH) by Kazhdan \etal \cite{Kazhdan:2003:RISH}, the LightField descriptor (LFD) by Chen \etal \cite{Chen:2003:ovsb}, and Fisher vectors extracted on the same rendered views of the shapes used as input to our networks. 

\begin{figure}[t]
\begin{center}
   \includegraphics[width=\linewidth,trim=-15pt 0pt -35pt 0pt,clip=true]{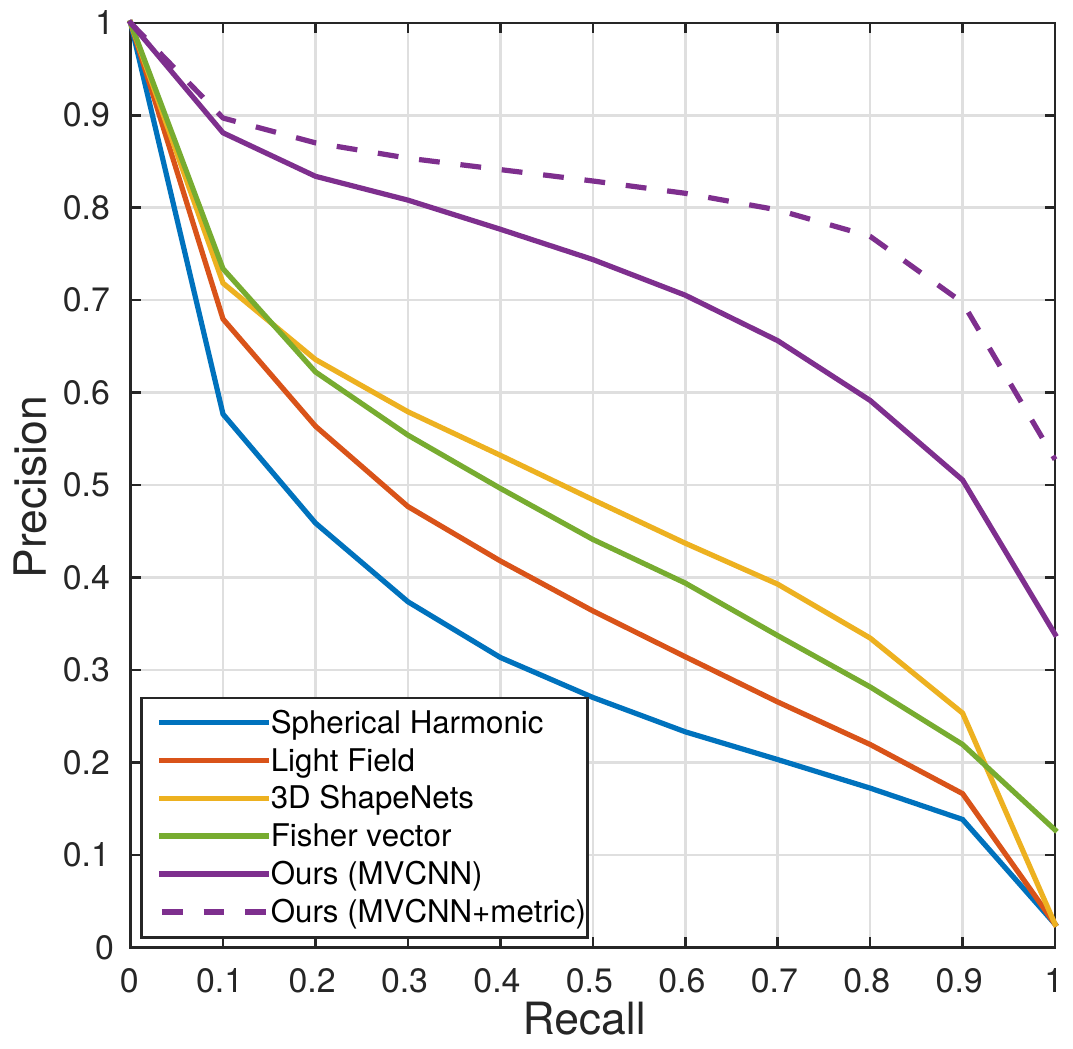}
\end{center}
\vskip -3mm
   \caption{Precision-recall curves for various methods for 3D shape retrieval on the ModelNet40 dataset. Our method significantly outperforms the state-of-the-art on this task achieving 80.2\% mAP.}
\label{fig:pr}
\vskip -3mm
\end{figure}

Results on shape classification and retrieval are summarized in Tab.~\ref{table:modelnet}.  Precision-recall curves are provided in Fig.~\ref{fig:pr}. Remarkably the Fisher vector baseline with just a single view achieves a classification accuracy of 78.8\% outperforming the state-of-the-art learned 3D descriptors (77.3\%~\cite{Wu:2015:ModelNet}). When all  12 views of the shape are available at test time (based on our first camera
setup), we can also average the predictions over these views. Averaging increases the performance of Fisher vectors to 84.8\%. 
The performance of Fisher vectors further supports our claim that 3D objects can be effectively represented using view-based 2D representations. The trends in performance for shape retrieval are similar.

Using our CNN baseline trained on ImageNet in turn outperforms Fisher vectors by a significant margin. Fine-tuning the CNN on the rendered views of the training shapes of ModelNet40 further improves the performance. By using all 12 views of the shape, its classification accuracy reaches 88.6\%, and mean average precision (mAP) for retrieval is also improved to 62.8\%. 

Our MVCNN outperforms all state-of-the-art descriptors as well as the Fisher vector and CNN baselines. With fine-tuning on the ModelNet40 training set, our model achieves 89.9\% classification accuracy, and 70.1\% mAP on retrieval using the first camera setup. If we do not make use of the assumption
about consistent upright orientation of shapes (second camera setup), the performance remains still intact, achieving 90.1\% classification accuracy and 70.4\% retrieval mAP. MVCNN constitutes an absolute gain of 12.8\% in classification accuracy compared to the state-of-the-art learned 3D shape descriptor \cite{Wu:2015:ModelNet} (77.3\% $\rightarrow$ 90.1\%). Similarly, retrieval mAP is improved by 21.2\% (49.2\% $\rightarrow$ 70.4\%). Finally, learning a low-rank Mahalanobis metric improves retrieval mAP further while classification accuracy remains almost unchanged, and the resulting shape descriptors become more compact ($d=4096, p=128$). 

We considered different locations to place the view-pooling layer in the MVCNN. 
Performance is not very sensitive among the later few layers (conv$_4$$\mathtt{\sim}$fc$_7$); however any location prior to conv$_4$ decreases classification accuracies significantly. We find conv$_5$ offers slightly better accuracies ($\mathtt{\sim}1\%$), and thus use it for all our experiments.  

\vskip -2mm
\paragraph{Saliency map among views.} 

For each 3D shape $S$, our multi-view representation consists of a set of $K$ 2D views $\{I_1,I_2 \dots I_K\}$. We would like to rank pixels in the 2D views \wrt their influence on the output score $F_c$ of the network (\eg taken from fc$_8$ layer) for its ground truth class $c$. Following \cite{Simonyan:2013:DIC}, saliency maps can be defined as the derivatives of $F_c$ \wrt the 2D views of the shape: 

\begin{equation}
\label{eq:saliency_supp}
\left[w_1,w_2 \dots w_K\right] = \left[\left.\frac{\partial F_c}{\partial I_1}\right|_{S}, \left.\frac{\partial F_c}{\partial I_2}\right|_{S} \dots \left.\frac{\partial F_c}{\partial I_K}\right|_{S}\right] 
\end{equation}

\begin{figure*}[t]
\begin{center}
   \includegraphics[width=\linewidth,trim=0pt 333pt 0pt 0pt,clip=true]{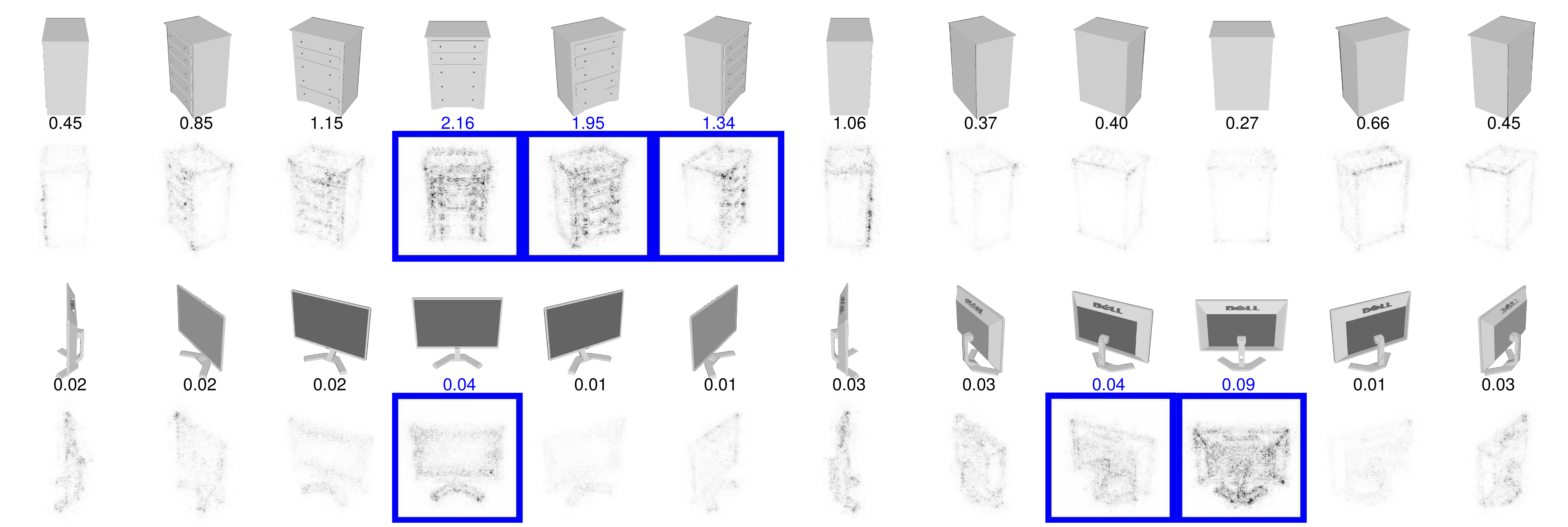}
\end{center}
\vskip -3mm
   \caption{Top three views with the highest saliency are highlighted in blue and the relative magnitude of gradient energy for each view is shown on top. The saliency maps are computed by back-propagating the gradients of the class score onto the images via the view-pooling layer. Notice that the handles of the dresser are the most discriminative features. (Figures are enhanced for visibility.) }
\vskip -3mm
\label{fig:saliency}
\end{figure*}

For MVCNN, $w$ in Eq.~\ref{eq:saliency_supp} can be computed using back-propagation with all the network parameters fixed, and can then be rearranged to form saliency maps for individual views. Examples of saliency maps are shown in Fig.~\ref{fig:saliency}. 

\subsection{Sketch Recognition: Jittering Revisited}
\label{sec:sketch}

Given the success of our aggregated descriptors on multiple views of a 3D object, 
it is logical to ask whether aggregating multiple views of a 2D
image could also improve performance. Here we show that this is indeed the case by 
exploring its connection with data jittering in the context of sketch recognition. 

Data jittering, or data augmentation, is a method to generate extra samples 
from a given image. It is the process of perturbing the image by transformations that 
change its appearance while leaving the high-level information (class label, attributes, \etc ) 
intact. Jittering can be applied at training time to augment training samples and to 
reduce overfitting, or at test time to provide more robust predictions. In particular, 
several authors \cite{Krizhevsky:2012:ICD,Chatfield:2014:RDD,DBLP:journals/corr/SzegedyLJSRAEVR14} 
have used data jittering to improve the performance of deep representations on 2D 
image classification tasks. 
In these applications, jittering at training time usually includes random image translations 
(implemented as random crops), horizontal reflections, and color perturbations. At test time jittering usually only includes a few crops (\eg, four at the corners, 
one at the center and their horizontal reflections). We now examine whether we can 
get more benefit out of jittered views of an image by using the same feature aggregation 
scheme we developed for recognizing 3D shapes.

The human sketch dataset \cite{Eitz:2012:HDH} contains 20,000 hand-drawn sketches of 250 object
categories such as airplanes, apples, bridges, \etc The accuracy of humans in recognizing 
these hand-drawings is only 73\% because of the low quality of some sketches. 
Schneider and Tuytelaars \cite{Schneider:2014:SCC} cleaned up the 
dataset by removing instances and categories that humans find hard to recognize. 
This cleaned dataset (SketchClean) contains 160 categories, 
on which humans can achieve 93\% recognition accuracy. Using SIFT Fisher vectors with spatial 
pyramid pooling and linear SVMs, Schneider and Tuytelaars achieved 68.9\% recognition accuracy on 
the original dataset and 79.0\% on the SketchClean dataset. We split the SketchClean dataset 
randomly into training, validation and test set,\footnote{The dataset does not come with a standard 
training/val/test split.} and report classification accuracy on the test set in 
Tab.~\ref{table:sketch-classification}. 

With an off-the-shelf CNN (VGG-M from \cite{Chatfield:2014:RDD}), we are able to get 77.3\% classification 
accuracy without any network fine-tuning. With fine-tuning on the training set, the 
accuracy can be further improved to 84.0\%, significantly surpassing the Fisher vector approach. 
These numbers are achieved by using the penultimate layer (fc$_7$) in the network as image 
descriptors and linear SVMs. 

Although it is impractical to get multiple views from 2D images, we can use jittering  
to mimic the effect of views. For each hand-drawn sketch, we do in-plane rotation with three angles: 
$-45\degree$, $0\degree$, $45\degree$, and also horizontal reflections (hence 6 
samples per image). We apply the two CNN variants (regular CNN and MVCNN) discussed earlier for aggregating multiple views of 3D shapes, and get 85.5\% (CNN w/o view-pooling) and 86.3\% (MVCNN w/ view-pooling on fc$_7$) classification accuracy respectively. The latter also has the advantage of a single, more compact descriptor for each hand-drawn sketch. 

With a deeper network architecture (VGG-VD, a network with 16 weight layers from \cite{DBLP:journals/corr/SimonyanZ14a}), we  achieve 87.2\% accuracy, further advancing the classification performance, and closely approaching human performance. 

\begin{table}
\setlength{\tabcolsep}{4pt}
\begin{center}
\begin{tabular}{lcc}
\toprule
\multicolumn{1}{m{4.4cm}}{\centering \textbf{Method}} & \multicolumn{1}{m{1cm}}{\centering \textbf{Aug.}} & \multicolumn{1}{m{1.6cm}}{\centering \textbf{Accuracy}} \\
\midrule
(1) FV \cite{Schneider:2014:SCC} & - &  79.0\%\\
\midrule
(2) CNN M & - & 77.3\% \\
(3) CNN M, fine-tuned & - & 84.0\%\\
(4) CNN M, fine-tuned &6$\times$ & 85.5\%\\
(5) MVCNN M, fine-tuned & 6$\times$ & 86.3\%\\
\midrule
(6) CNN VD & - & 69.3\% \\
(7) CNN VD, fine-tuned & - & 86.3\%\\
(8) CNN VD, fine-tuned & 6$\times$ & 86.0\%\\
(9) MVCNN VD, fine-tuned & 6$\times$& \textbf{87.2\%}\\
\midrule
(10) Human performance & n/a & 93.0\% \\
\bottomrule
\end{tabular}
\end{center}
\vskip -3mm
\caption{Classification results on SketchClean. Fine-tuned CNN models significantly outperform Fisher vectors \cite{Schneider:2014:SCC} by a significant margin. MVCNNs are better than CNN trained with data jittering. The results are shown with two different CNN architectures -- VGG-M (row 2-5) and VGG-VD (row 6-9).}
\label{table:sketch-classification}
\vskip -3mm
\end{table}

\subsection{Sketch-based 3D Shape Retrieval}
\label{sec:sketch-retrieval}

Due to the growing number of online 3D repositories, 
many approaches have been investigated to perform efficient 3D shape retrieval. 
Most online repositories
(\eg 3D Warehouse, TurboSquid, Shapeways) 
provide only text-based 
search engines or hierarchical catalogs for 3D shape retrieval.
However, it is hard to convey stylistic and geometric variations using only textual descriptions, thus sketch-based shape retrieval  \cite{Yoon:2010:SMR,Shao:2011:DSM,Eitz:2012:SSR} has been proposed as an alternative for users to retrieve shapes with an approximate sketch of the desired 3D shape in mind. 
Sketch-based retrieval is challenging since it involves two heterogeneous data domains (hand-drawn sketches and 3D shapes), and sketches can be highly abstract and visually different from 
target 3D shapes. Here we demonstrate the potential strength of MVCNNs in sketch-based shape retrieval. 

\begin{figure}[t]
\begin{center}
   \includegraphics[width=\linewidth,trim=0pt 0pt 0pt -40pt,clip=true]{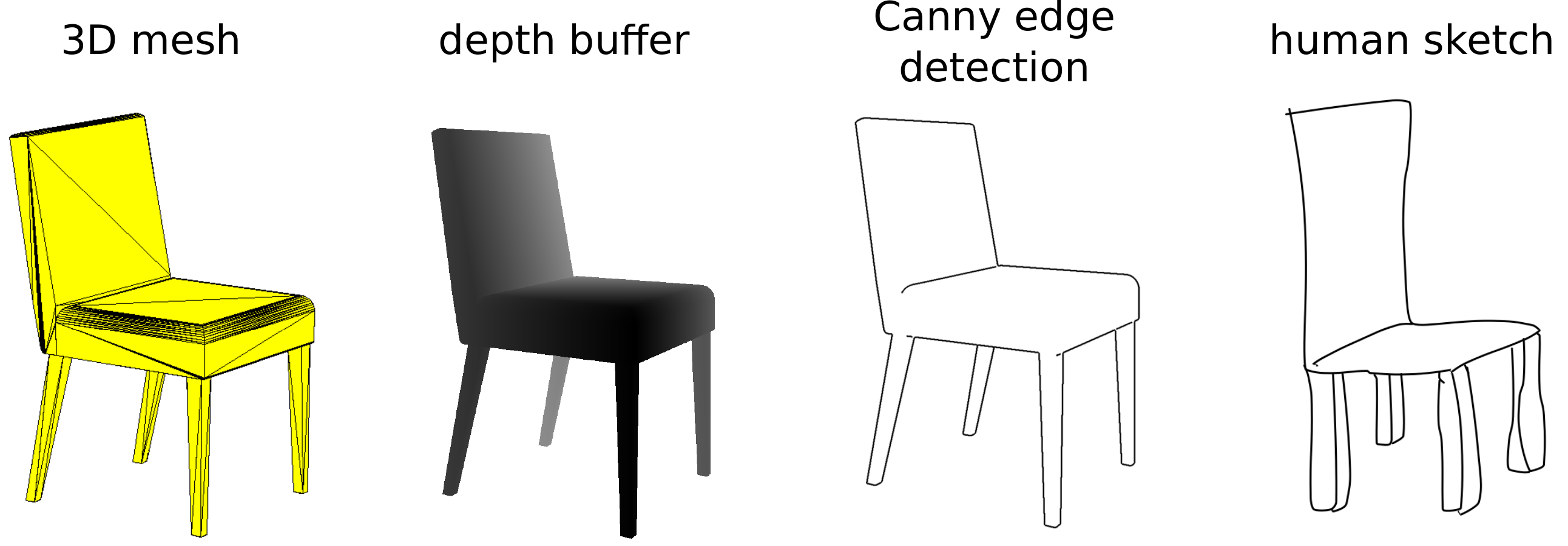}
\end{center}
\vskip -3mm
   \caption{Line-drawing style rendering from 3D shapes.}
\label{fig:sketch-based-retrieval}
\vskip -3mm
\end{figure}

\begin{figure}[t]
\begin{center}
   \includegraphics[width=\linewidth]{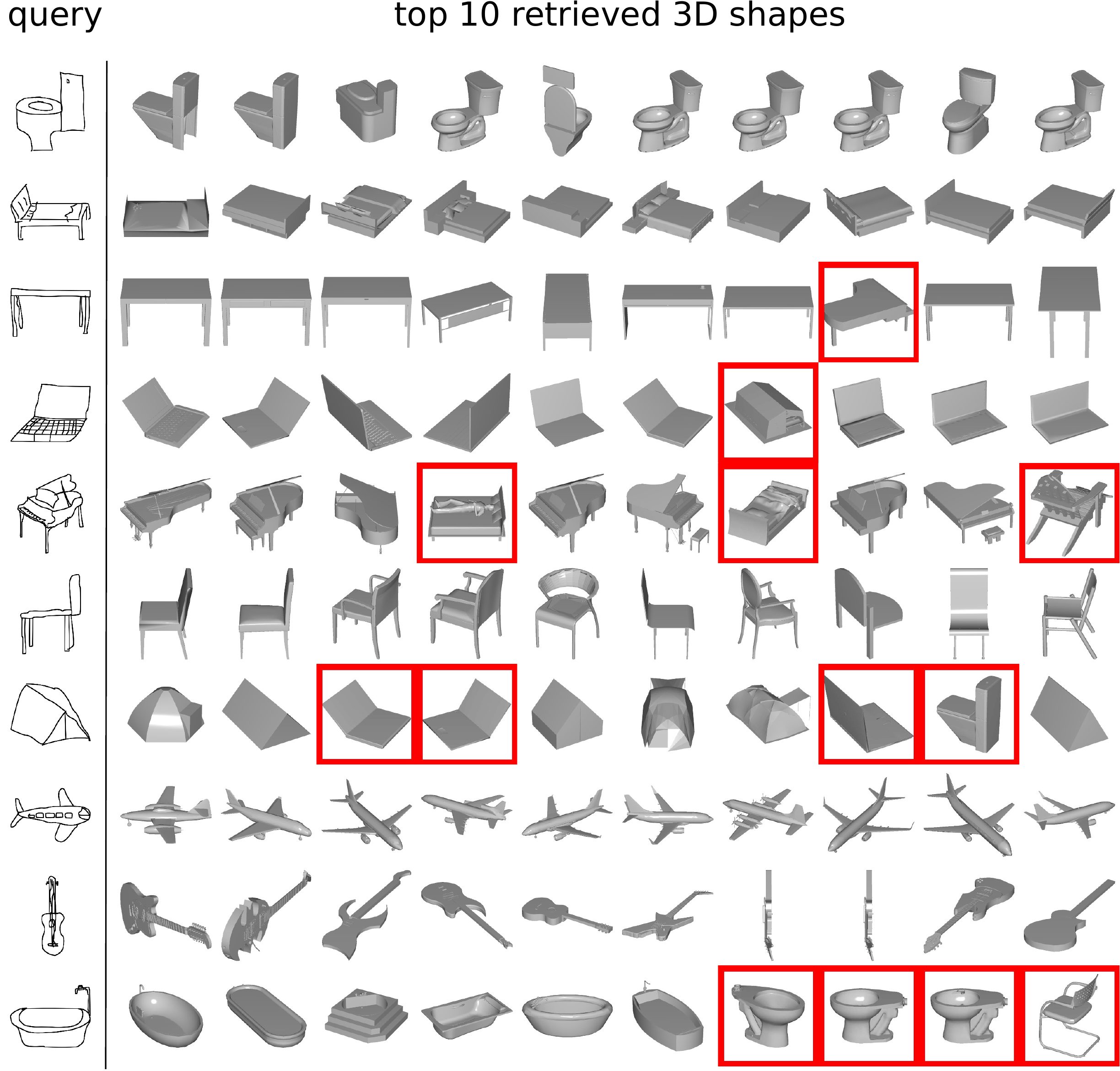}
\end{center}
\vskip -3mm
   \caption{Sketch-based 3D shape retrieval examples. Top matches are shown for each query, with mistakes highlighted in red.}
\label{fig:sketch-based-retrieval-results}
\vskip -3mm
\end{figure}

For this experiment, we construct a dataset containing 193 sketches and 790 CAD models from 10 categories existing in both SketchClean and ModelNet40. Following \cite{Eitz:2012:SSR}, we produce renderings of  3D shapes with a style similar to hand-drawn sketches (see Fig.~\ref{fig:sketch-based-retrieval}). 
This is achieved by detecting Canny edges on the depth buffer (also known as $z$-buffer) from 12 viewpoints. These edge maps are then passed through CNNs to obtain image descriptors. Descriptors are also extracted from 6 perturbed samples of each query sketch in the manner described in Sect.~\ref{sec:sketch}. Finally we rank 3D shapes \wrt ``average minimum distance" (Eq.~\ref{eq:avgmindist}) to the sketch descriptors. Representative retrieval results are shown in Fig.~\ref{fig:sketch-based-retrieval-results}.

We are able to retrieve 3D objects from the same class with the query sketch, as well as being visually similar, especially in the top few matches. Our performance is 36.1\% mAP on this dataset. Here we use the VGG-M network trained on ImageNet without any fine-tuning on either sketches or 3D shapes. With a fine-tuning procedure that optimizes a distance measure between hand-drawn sketches and 3D shapes, \eg, by using a Siamese Network \cite{Chopra:2005:LSM}, retrieval performance can be further improved. 

\section{Conclusion}
While the world is full of 3D shapes, as humans at least, we
understand that world is mostly through 2D images. We have shown that
using images of shapes as inputs to modern learning architectures, we
can achieve performance better than any previously published results,
including those that operate on direct 3D representations of shapes.

While even a n{\"a}ive usage of these multiple 2D projections yields
impressive 
discrimination performance, 
by building descriptors that are aggregations of
information from multiple views, we can achieve compactness,
efficiency, and better accuracy. 
In addition, by relating the content of 3D shapes to 2D representations like
sketches, we can retrieve these 3D shapes at high accuracy using sketches, and leverage the implicit knowledge of 3D shapes 
contained in their 2D views.

There are a number of directions to explore in future work. One is
to experiment with different combinations of 2D views. Which
views are most informative? How many views are necessary for a given
level of accuracy? Can informative views be selected on the fly?


Another obvious question is whether
our view aggregating techniques can be used for building compact and
discriminative descriptors for real-world 3D objects from multiple views, or
automatically from video, rather than merely for 3D polygon mesh models. 
Such investigations could be immediately
applicable to widely studied problems such as object recognition and face
recognition.

\paragraph{Acknowledgements}

We  thank  Yanjie Li for her help on rendering meshes. We thank NVIDIA for their generous donation of  GPUs used in this research. Our work was partially supported
 by NSF (CHS-1422441).

\newpage
\appendix

{\raggedleft{} \bf \Large Appendix}
\vspace{0.3cm}

Here we provide additional evaluations and visulizations of our multi-view CNN (MVCNN), including 
\begin{inparaenum}[\itshape a\upshape)]
\item confusion matrix of 3D shape classification;
\item additional view-based saliency maps; and 
\item examples of correctly and wrongly classified hand-drawn sketches. 
\end{inparaenum} 

\section{3D shape classification}
\begin{figure}[tbh]
\begin{center}
   \includegraphics[width=1.025\linewidth,trim=38 40 62 20mm,clip=true]{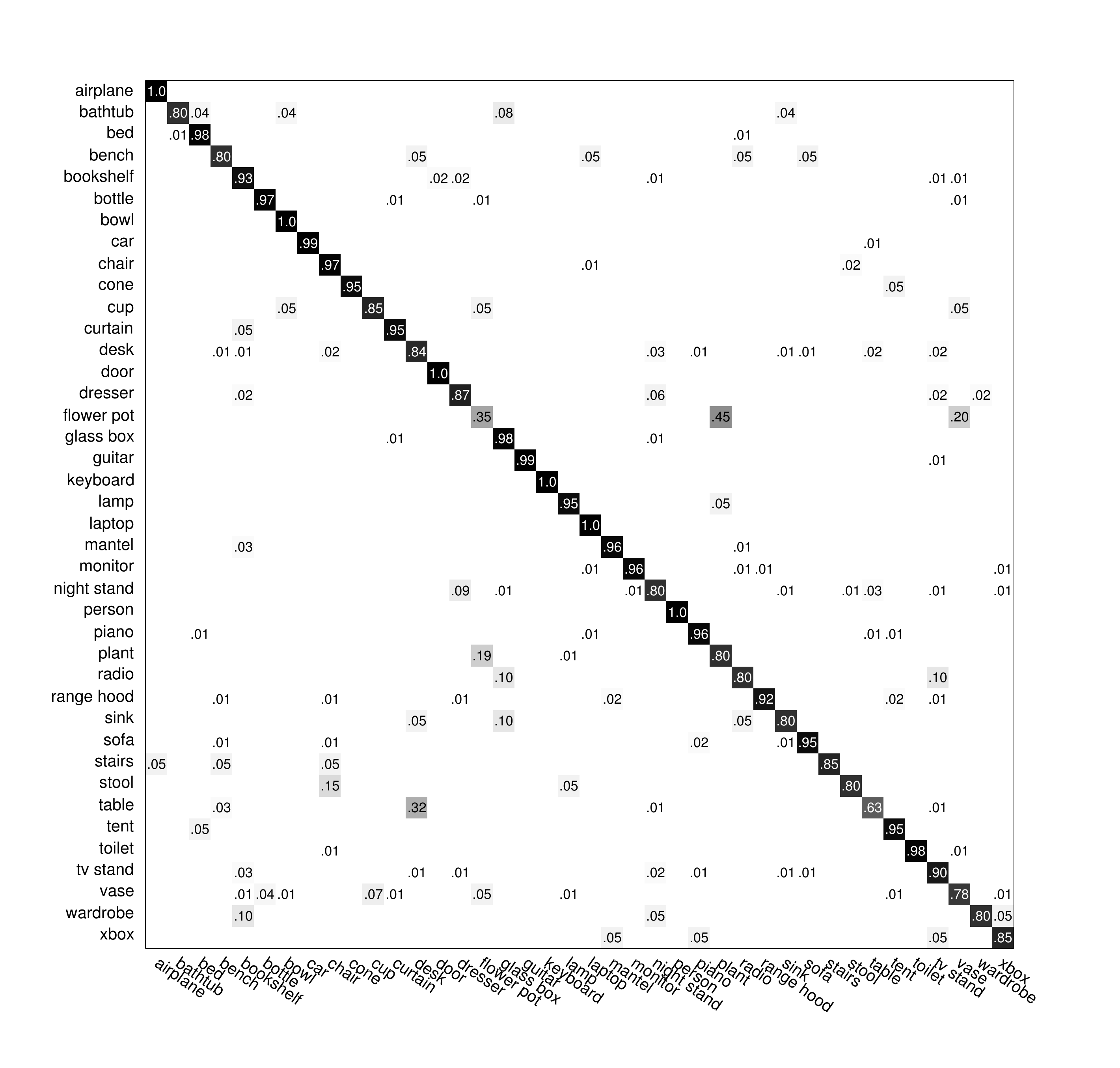}
\end{center}
   \caption{Confusion matrix of ModelNet40 classification. } 
\label{fig:conf_supp}
\end{figure}

Confusion matrix of 3D shape classification on ModelNet40 is given in Figure~\ref{fig:conf_supp}. 
Here MVCNN with fine-tuning on 12 views (row 11 in Table 1 of main submission) is used.

Top confusions occur at 
\begin{inparaenum}[1\upshape)]
\item \emph{flower pot} $\rightarrow$ \emph{plant} (45\%), 
\item \emph{table} $\rightarrow$ \emph{desk} (32\%), 
\item \emph{flower pot} $\rightarrow$ \emph{vase} (20\%), 
\item \emph{plant} $\rightarrow$ \emph{flower} (19\%), and
\item \emph{stool} $\rightarrow$ \emph{chair} (15\%). 
\end{inparaenum} 
Distinctions between some of these pairs are ambiguous even for humans.  

\section{Image-specific class saliency visualization across views}
\begin{figure*}[t]
\begin{center}
   \includegraphics[width=\linewidth,trim=0 0 0 18mm,clip=true]{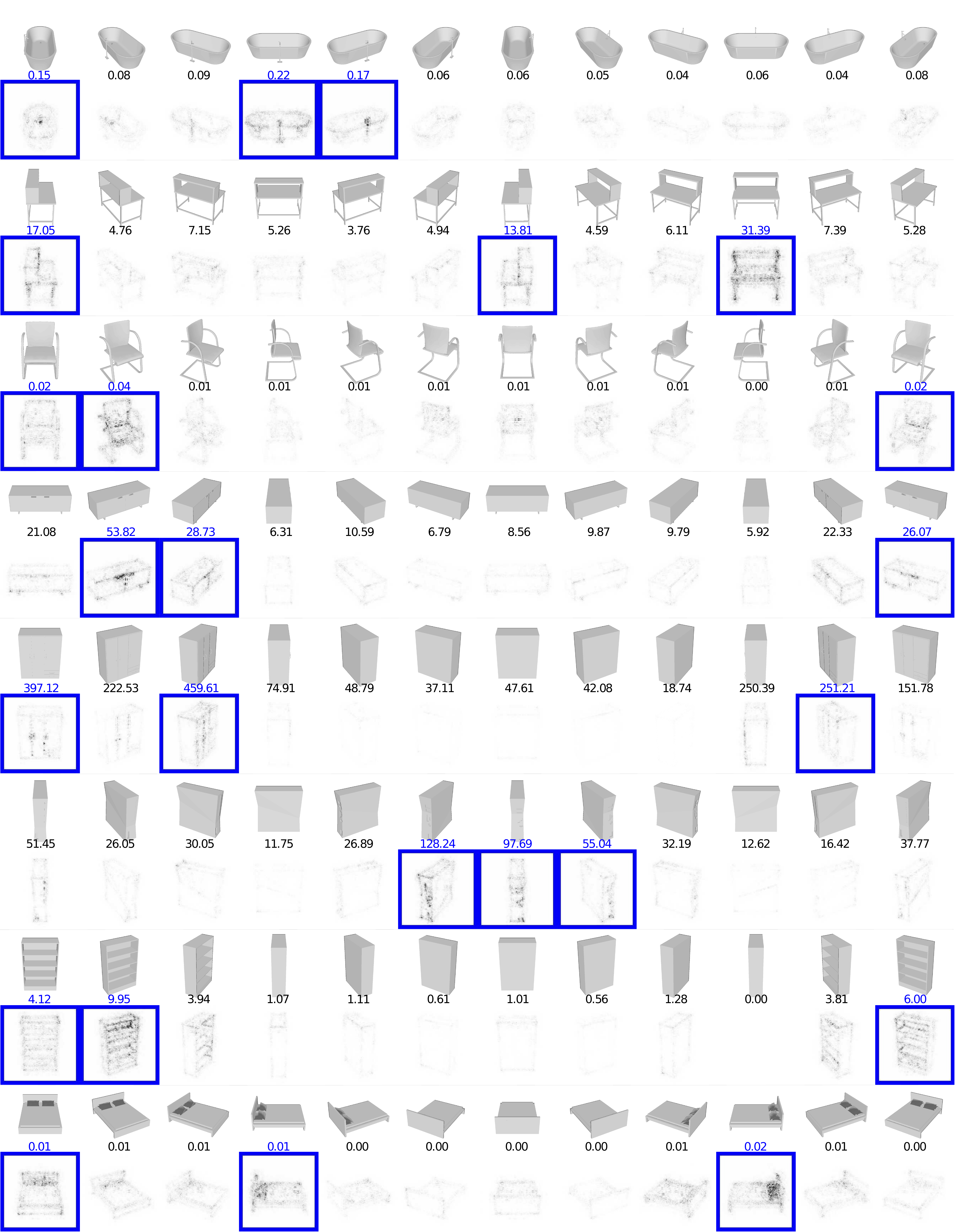}
\end{center}
   \caption{Additional examples of saliency maps. Top three views with the highest saliency are highlighted in blue and the relative magnitudes of gradient energy for each view is shown on top. }
\label{fig:saliency_supp}
\end{figure*}

Additional examples of saliency maps are shown in Figure~\ref{fig:saliency_supp}. Note that the saliency maps tend to highlight
\begin{inparaenum}[\itshape a\upshape)]
\item the most canonical views accross views, \eg the front view of the bench; and 
\item the most discriminative parts within views, \eg the faucet and the sink hole of the bathtub. 
\end{inparaenum}

\section{Sketch classification}
\begin{figure*}[tbh]
\begin{center}
   \includegraphics[width=\linewidth]{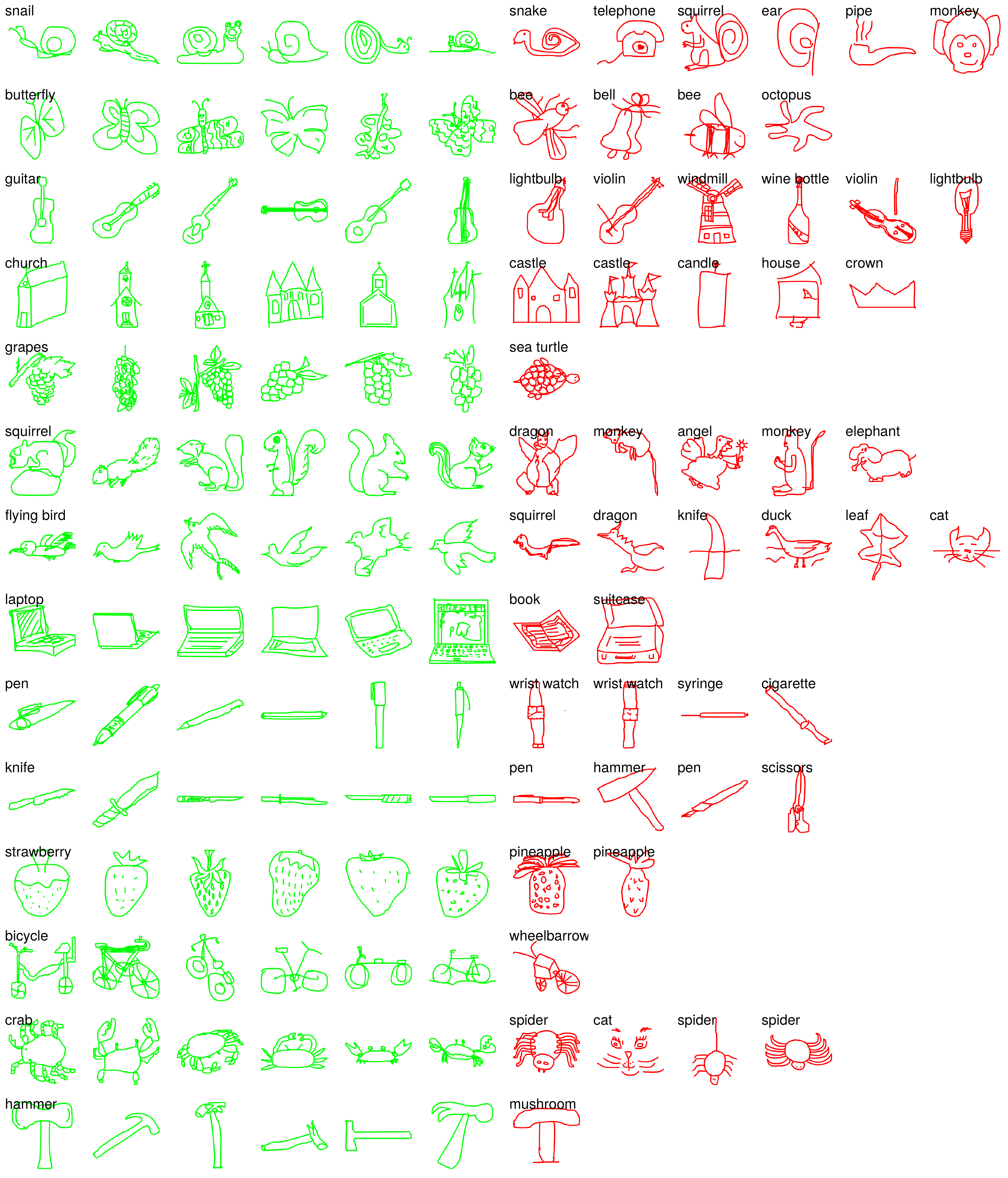}
\end{center}
   \caption{Examples of correctly and wrongly classified hand-drawn sketches. All sketches in each row are classified into the class labeled on top left. False positives are in red, with their ground truth 
classes labeled on top.} 
\label{fig:sketch_supp}
\end{figure*}

Examples of correctly and wrongly classified hand-drawn sketches are shown in Figure~\ref{fig:sketch_supp}. Most misclassified sketches contain visually similar components with the target class, 
\eg spider and crab have a similar layout of legs, and some are difficult to recognize even for humans. 

\section{Document changelog}
\noindent \textbf{v1} Initial version.

\noindent \textbf{v2} An updated ModelNet40 training/test split is used for experiments 
in order to be consistent with \cite{Wu:2015:ModelNet}. Performance of most methods drops 
a bit because of the smaller training set (the full ModelNet40 was used in \textbf{v1}). 
Results with low-rank Mahalanobis metric learning are added. 

\noindent \textbf{v3} A second camera setup without the upright orientation assumption is added. Some accuracy and mAP numbers are changed slightly because a small issue in mesh rendering related to specularities is fixed.

{\small
\bibliographystyle{ieee}
\bibliography{shapes_arxiv}
}

\end{document}